\documentclass[a4paper,conference]{IEEEtran}
\ifCLASSINFOpdf
\else
\fi
\usepackage{graphicx}
\usepackage{amsfonts}
\usepackage{graphicx}
\usepackage{epstopdf}
\usepackage{epsfig}
\usepackage{multirow}
\usepackage{array}
\usepackage{hhline}
\usepackage{makecell}
\usepackage{cite}
\usepackage{amsmath}
\usepackage{algorithmic}
\usepackage{algorithm}
\usepackage{xcolor}
\begin{document}

\title{Semi-Supervised Domain Adaptation via Selective Pseudo Labeling and Progressive Self-Training}

%\title{Selective Pseudo Labeling and Progressive Self-Training for Semi-Supervised Domain Adaptation}

%\title{Progressive Self-Training with Selective Pseudo Labeling for Semi-Supervised Domain Adaptation}

%\title{Selective Self-Training with Progressive Psuedo Label Refinery for Semi-Supervised Domain Adaptation}

%\title{Selective Pseudo Labeling and Label Noise-Robust Learning for Semi-Supervised Domain Adapation}

\author{\IEEEauthorblockN{Yoonhyung Kim}
\IEEEauthorblockA{School of Electrical Engineering\\
Korea Advanced Institute of\\
Science and Technology (KAIST)\\
291 Daehak-ro, Yuseong-gu, Daejeon,\\
34141, Republic of Korea\\
Email: yhkim1127@kaist.ac.kr}
\and
\IEEEauthorblockN{Changick Kim}
\IEEEauthorblockA{School of Electrical Engineering\\
Korea Advanced Institute of\\
Science and Technology (KAIST)\\
291 Daehak-ro, Yuseong-gu, Daejeon,\\
34141, Republic of Korea\\
Email: changick@kaist.ac.kr}
}


%Telephone: (800) 555--1212, Fax: (888) 555--1212}


\maketitle

\begin{abstract}
Domain adaptation (DA) is a representation learning methodology that transfers knowledge from a label-sufficient source domain to a label-scarce target domain. While most of early methods are focused on unsupervised DA (UDA), several studies on semi-supervised DA (SSDA) are recently suggested. In SSDA, a small number of labeled target images are given for training, and the effectiveness of those data is demonstrated by the previous studies. However, the previous SSDA approaches solely adopt those data for embedding ordinary supervised losses, overlooking the potential usefulness of the few yet informative clues. Based on this observation, in this paper, we propose a novel method that further exploits the labeled target images for SSDA. Specifically, we utilize labeled target images to selectively generate pseudo labels for unlabeled target images. In addition, based on the observation that pseudo labels are inevitably noisy, we apply a label noise-robust learning scheme, which progressively updates the network and the set of pseudo labels by turns. Extensive experimental results show that our proposed method outperforms other previous state-of-the-art SSDA methods.
\end{abstract}


% \fi
\IEEEpeerreviewmaketitle

\section{Introduction}
\label{sec:intro}
\begin{figure}[t!]
\begin{center}
\begin{minipage}{1.0\linewidth}
\centering{\epsfig{file=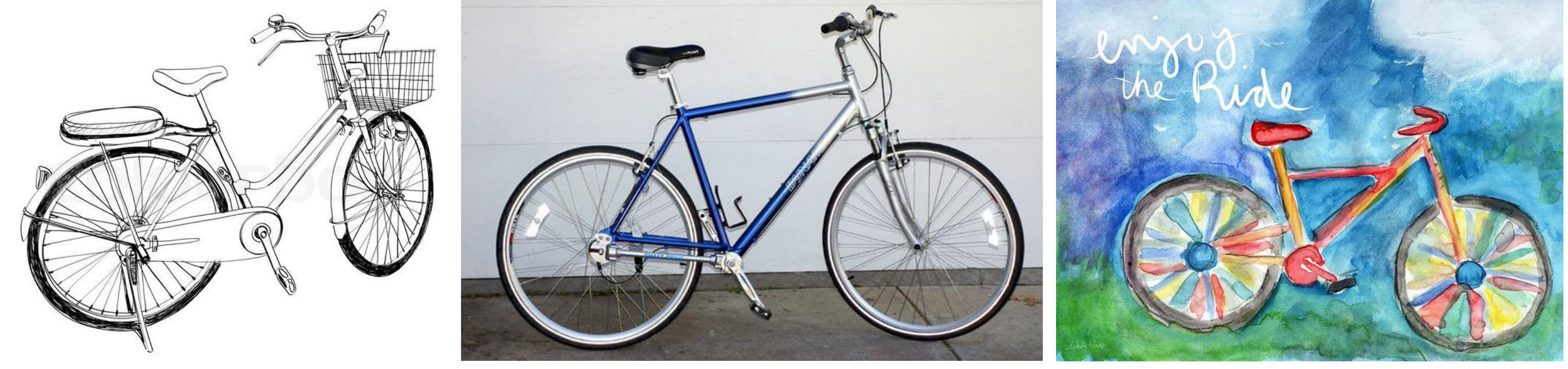,width=1.0\linewidth}}
\end{minipage}
\end{center}
\caption{A set of images in LSDAC dataset\cite{peng2019moment}  to illustrate the notion of domain shift. The above images are examples in Sketch, Real, and Painting domains, respectively.}
\label{fig1}
\end{figure}

When encountered an image representing a single object, humans can easily recognize its identity regardless of domain characteristics. For example, we can instantly figure out that all images in Fig. \ref{fig1} represent a ``bicycle'' even though there obviously exists contextual disparity (or domain shift\cite{pan2009survey}) among the images. Meanwhile, deep neural networks trained on a single domain are known to be fragile to the domain shift due to the strong dependency upon training data. One simple yet naive solution is to prepare a large amount of training data for each domain, but tremendous expenses are compelled as well. In addition, tagging a label for every image in the target domain is particularly costly and time-consuming if the number of classes becomes larger. To overcome this problem, various representation learning approaches named domain adaptation (DA) have been proposed in recent years\cite{wang2018deep}.

\begin{figure*}[t!]
\begin{center}
\begin{minipage}{1.0\linewidth}
\centering{\epsfig{file=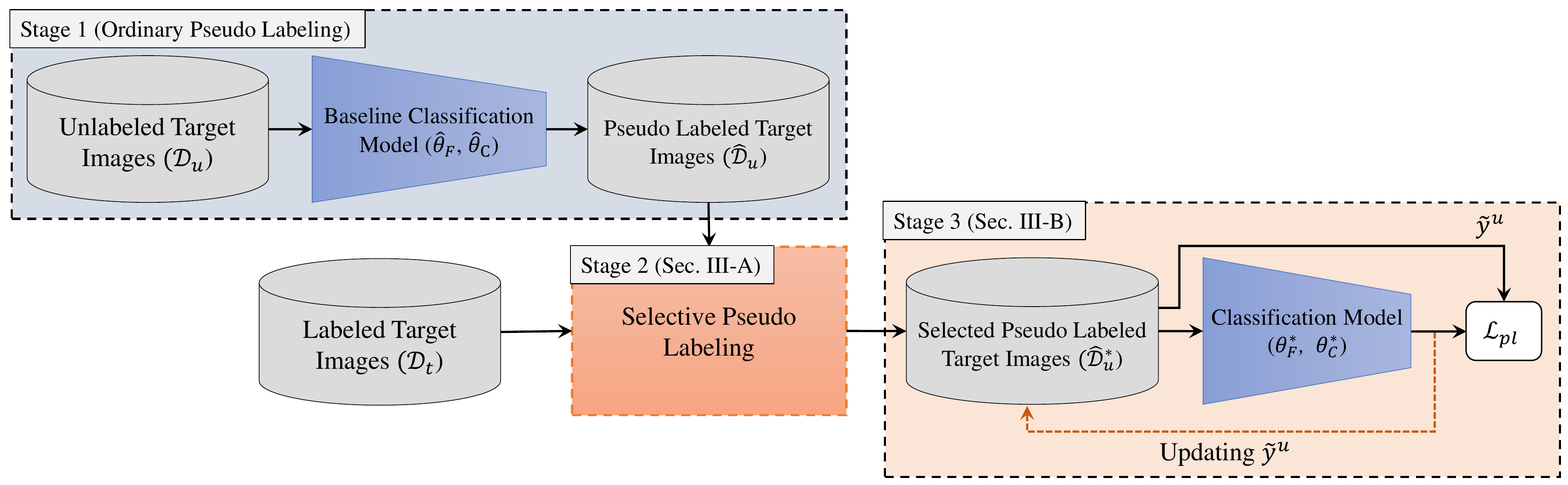,width=1.0\linewidth}}
\end{minipage}
\end{center}
\caption{The overall pipeline of the proposed SSDA method.}
\label{fig:proposed_pipeline}
\end{figure*}

The goal of DA is to enhance the performance of classifying images in a label-scarce domain (target domain) by leveraging knowledge of a label-sufficient domain (source domain). Majority of early methods \cite{ganin2015dann, long2018cdan, volpi2018gan, hu2018duplex, saito2018mcd, french2018se, kurmi2018attention, gong2019dlow, ma2019gcan, choi2019pseudo} are devoted to unsupervised domain adaptation (UDA), which assumes all target images are unlabeled while source images are fully labeled. Recently, a pioneering study \cite{saito2019mme} on semi-supervised domain adaptation is introduced, which assumes a few labeled target images are additionally given (e.g., one or three examples per each class). In the study, a few-shot feature embedding scheme \cite{chen2019closer} is incorporated to enhance the effectiveness of labeled target images. In addition, by means of the minimax entropy-based learning scheme, the method outperforms other UDA methods, which are trained with SSDA setups (i.e., additional supervisions on the few labeled target images). One of empirical discoveries reported in \cite{saito2019mme} is that training with additional labeled data in the target domain can considerably enhance the performance even though the quantity of those data is extremely small. This implies that the few labeled target images serve as critical clues to resolve SSDA problems. However, in spite of the significance of the labeled target images, their usage in the existing SSDA methods is limited to embedding  them into ordinary supervised losses, such as cross entropy loss.

In this paper, we propose a new SSDA method that exploits the labeled target images more actively by treating them as ‘golden’ samples for SSDA. To this end, we employ the few labeled target images for selectively assigning pseudo labels to unlabeled target images. Training with pseudo labels \cite{lee2013pseudo} requires careful treatments since incorrect pseudo labels may result in performance degradation. Our strategy to deal with pseudo labels is composed of two major components. First, to acquire pseudo labels with high reliability, we propose to select and utilize restricted amounts of pseudo labels based on an analysis in the feature space. Here, the basis of our reasoning is that deep features that lead to correct pseudo labels are usually clustered with those of labeled target images. Second, based on the observation that pseudo labels are inevitably noisy (i.e., containing incorrect labels), we propose to apply a label noise-robust learning scheme \cite{tanaka2018joint} that alternately updates pseudo labels and deep networks. By means of this alternate updating scheme, the network and the set of pseudo labels are progressively optimized. The overall pipeline of the proposed SSDA method is illustrated in Fig. \ref{fig:proposed_pipeline}. \color{black} Experimental results on LSDAC \cite{peng2019moment}, Office-Home \cite{venkateswara2017deep}, and Office \cite{saenko2010adapting} datasets demonstrate that our method outperforms other previous state-of-the-art methods.

The rest of this paper is organized as follows. In Section \ref{sec:related}, previous studies that are related to our work are introduced. In Section \ref{sec:proposed}, the details of our proposed method are explained. In Section \ref{sec:exp}, experimental setups and results are reported, and concluding remarks are given in Section \ref{sec:conclusion}.

\section{Related Work}
In this section, we review existing studies that are related to our work. First, we introduce previous domain adaptation methods for image classification. Second, we review learning schemes that are robust to noisy labels and clarify our strategy to apply those methods to SSDA.

\label{sec:related}
\subsection{Domain Adaptation for Image Classification}
Existing domain adaptation methods for image classification can be categorized into unsupervised and semi-supervised domain adaptation approaches. Both approaches consider the case that source and target domains share the same set of image categories, whereas the quantity of labels in the target domain is much smaller than that in the source domain.

Most of early studies are focused on UDA, which assumes that all images in the target domain are unlabeled. As a pioneering method for UDA, Ganin and Lempitsky \cite{ganin2015dann} propose an adversarial learning approach to aligning feature distributions of source and target domains. Through the adversarial learning, the feature extractor is trained to deceive the domain classifier by making features of the target domain be indistinguishable from those of the source domain. The adversarial learning process is implemented by inserting the gradient reversal layer (GRL) between the feature extractor and the domain classifier. This adversarial learning mechanism is widely adapted to other UDA approaches \cite{long2018cdan, saito2018mcd, french2018se, kurmi2018attention, ma2019gcan, choi2019pseudo} to aligning feature spaces. Different from those feature-level adaptation approaches, there are several pixel-level adaptation approaches that augment the scales of training sets by transferring images across the two domains \cite{volpi2018gan, hu2018duplex, gong2019dlow}. A common limitation of UDA methods is that the adaptation performance is severely degraded for adaptation scenarios involving a large domain shift. This is due to the harsh experimental setups of UDA that target labels are not given at all.

Recently, to address the domain adaptation problem in a more practical and realistic way, SSDA methods received a great attention. Unlike the UDA schemes, SSDA assumes that a few target labels (e.g., one or three examples for each class) are additionally given for domain adaptation. As a pioneering approach for SSDA, Saito et al. \cite{saito2019mme} propose a minimax entropy-based method. In the study, the few-shot feature embedding scheme \cite{chen2019closer} and the minimax entropy-based learning schemes are incorporated for SSDA. The empirical results in \cite{saito2019mme} show that additional supervisions on the few labeled target images can fairly increase the performance of domain adaptation methods \cite{saito2019mme, saito2017adversarial, ganin2015dann, long2018cdan, grandvalet2005semi}, implying the importance of those data. However, in spite of the significance of labeled target images, the use of those data is restricted to embedding ordinary supervised losses. Unlike those previous methods, in this paper, we propose to further utilize the labeled target images to select reliable pseudo labels for unlabeled target images. Training deep neural networks with pseudo labels \cite{lee2013pseudo} is one of the self-training mechanisms, and it requires careful treatments since incorrect pseudo labels can severely degrade the performance. To figure out pseudo labels with high reliability, we conduct feature analysis by exploiting both the labeled and the unlabeled target images. The details of this process are explained in Sec. \ref{sec:2A}.

\begin{figure*}[h!]
\begin{center}
\begin{minipage}{1.0\linewidth}
\centering{\epsfig{file=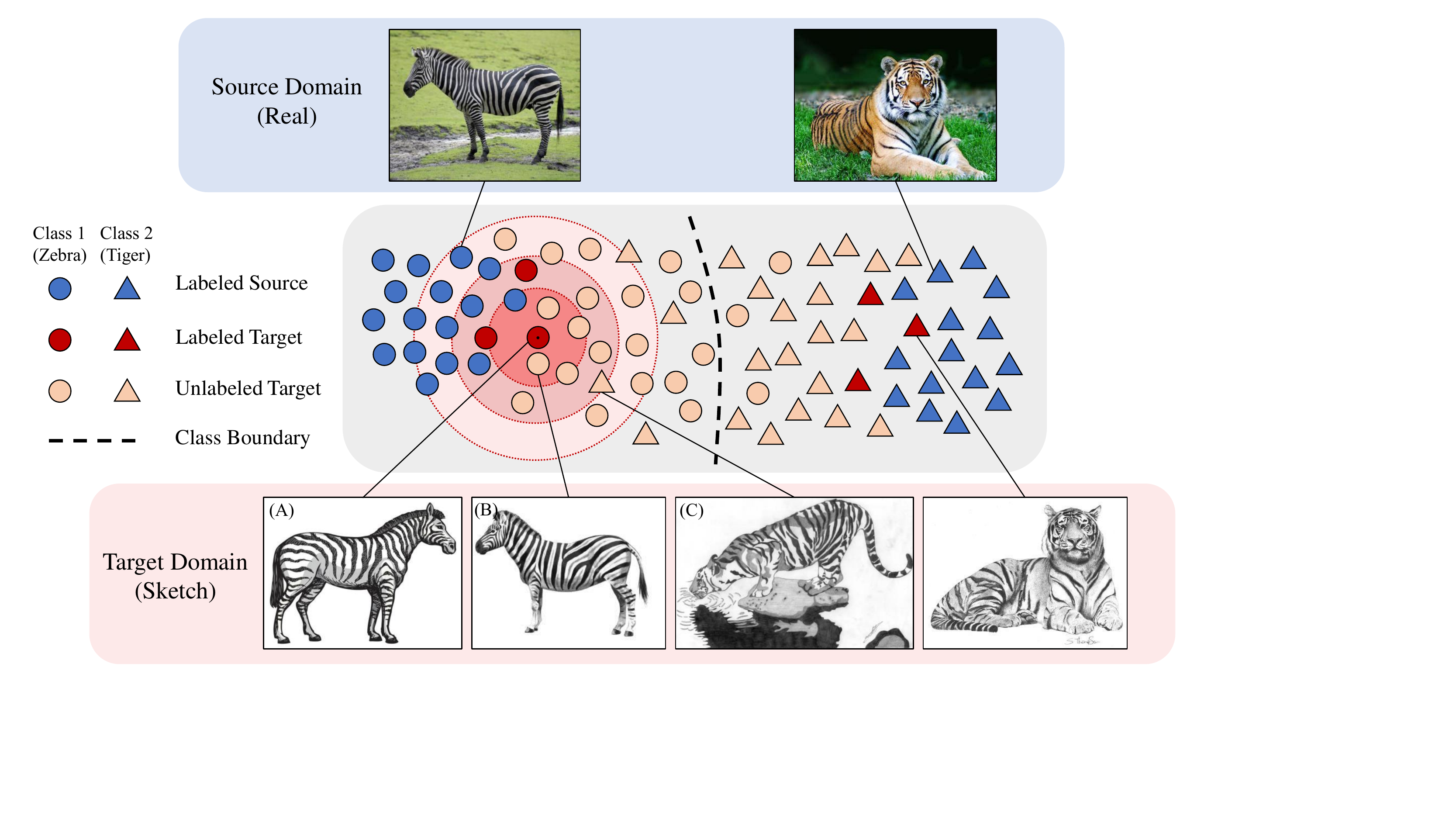,width=1.0\linewidth}}
\end{minipage}
\end{center}
\caption{A toy example to illustrate our motivation of the selective pseudo labeling approach. Generally, in the feature space, labeled source and target features are well assorted by the class boundary, whereas unlabeled target features are not. Thus, assigning pseudo labels to all unlabeled samples may generate numerous incorrect pseudo labels. Our motivation of selecting reliable pseudo labels is based on the observation that unlabeled target features leading to correct pseudo labels are located close to labeled target features in the feature space. For instance, in the figure, (B) is a correctly classified example whereas (C) is an incorrectly classified example. The feature distance between (A) and (B) is 0.91 and the feature distance between (A) and (C) is 1.64 (ResNet-34). By selectively assigning pseudo labels to unlabeled target images with relatively small feature distances, we can enhance the reliability of the set of pseudo labels. The example images are from the LSDAC dataset \cite{peng2019moment}. Best viewed in color.}
\label{fig:proposed1}
\end{figure*}

\subsection{Learning with Noisy Labels}
Training deep neural networks requires large-scale datasets, which are composed of images and corresponding label annotations. However, collecting clean labels for large-scale datasets is costly, and in practice there often exist noisy labels. By `noisy', we mean the labels may contain incorrect annotations, and learning with noisy labels is a challenging issue that is recently addressed by numerous studies \cite{tanaka2018joint, kim2019nlnl, ghosh2017robust, northcutt2017learning, zhang2018generalized, vahdat2017toward}. There are various existing approaches for learning with noisy labels, such as embedding label noise-robust loss functions \cite{ghosh2017robust, zhang2018generalized}, applying the joint optimization framework \cite{tanaka2018joint}, and filtering out noisy labels \cite{kim2019nlnl, northcutt2017learning}. Those methods are verified on image classification datasets, which contain intentionally generated noisy labels.

Our motivation of adapting the label noise-robust learning scheme to SSDA is derived from the fact that pseudo labels are inevitably noisy. To enhance the performance of the network trained on pseudo labels, we incorporate the joint optimization framework \cite{tanaka2018joint}, which is demonstrated to be robust to noisy labels of large-scale datasets. The key idea of the framework is to progressively update the network and the set of noisy labels by turns, pursuing positive interactions between the two components. The detailed descriptions of our label noise-robust learning scheme, which is motivated by \cite{tanaka2018joint}, is introduced in Sec. \ref{sec:2B}. To the best of our knowledge, this is the first trial to adapt the label noise-robust learning scheme to self-training with pseudo labels.

\begin{figure*}[t!]
\begin{center}
\begin{minipage}{1.0\linewidth}
\centering{\epsfig{file=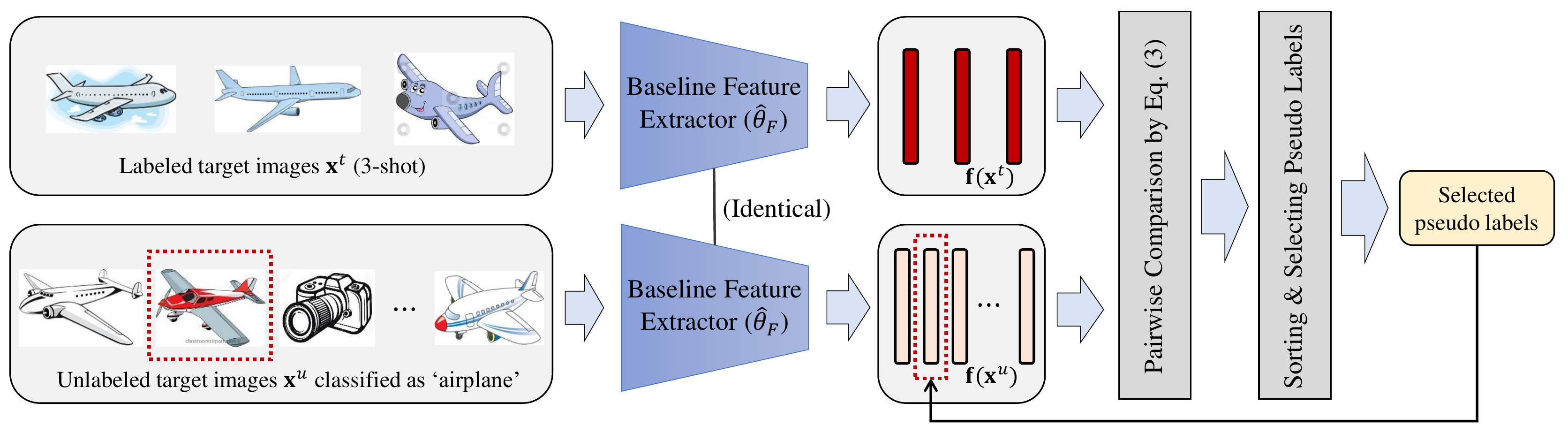,width=1.0\linewidth}}
\end{minipage}
\end{center}
\caption{An overview of the proposed selective pseudo labeling pipeline, which is explained in Sec. \ref{sec:2A}. The above procedure is conducted for each class in the target domain. The above figure illustrates the example of the `airplane' class in the Clipart domain of the LSDAC dataset\cite{peng2019moment}. Best viewed in color.}
\label{fig:proposed2}
\end{figure*}

\section{Proposed Method}
\label{sec:proposed}
The goal of semi-supervised domain adaptation is to train a classification model that is oriented to a target domain by using image sets in both domains. In the source domain, we are given source images and the corresponding labels $\mathcal{D}_{s}=\{(\mathbf{x}_{i}^{s}, {y}_{i}^{s})\}_{i=1}^{n_{s}}$. In the target domain, unlabeled images $\mathcal{D}_{u}=\{\mathbf{x}_{i}^{u}\}_{i=1}^{n_{u}}$ and a small number of labeled images $\mathcal{D}_{t}=\{(\mathbf{x}_{i}^{t}, {y}_{i}^{t})\}_{i=1}^{n_{t}}$ are given. In SSDA, the classification model is trained on $\mathcal{D}_{s}$, $\mathcal{D}_{u}$, $\mathcal{D}_{t}$ and tested on $\mathcal{D}_{u}$. The classification model is composed of a feature extractor $F(\cdot; \theta_{F})$ and a classifier $C(\cdot ; \theta_{C})$, where $\theta_{F}$ and $\theta_{C}$ are weight vectors of the feature extractor and the classifier, respectively. For an input image $\mathbf{x}$, its feature vector and output prediction encoded by the model are denoted as $\mathbf{f(x)}$ and $\mathbf{p(x)}$, respectively. Thus, $\mathbf{p(x)}=C(\mathbf{f(x)}; \theta_{C})=C(F(\mathbf{x}; \theta_{F}); \theta_{C})$.

As illustrated in Fig. \ref{fig:proposed_pipeline}, our proposed method is composed of three stages. The first stage is to train a baseline model to generate pseudo labels. In this paper, we adopt the minimax entropy-based approach \cite{saito2019mme} to train the baseline models for all experiments. The weight vectors of the feature extractor and the classifier of the trained baseline model are denoted as $\hat{\theta}_{F}$ and $\hat{\theta}_{C}$, respectively. The next two stages of our proposed method are explained in the following two subsections.

\subsection{Selective Pseudo Labeling Approach}
\label{sec:2A}
By using the baseline model that is acquired in the previous stage, we apply forward pass operations to unlabeled target images to obtain $\hat{\mathcal{D}}_{u}=\{(\mathbf{x}_{i}^{u}, \widetilde{\mathbf{y}}_{i}^{u}, \hat{y}_{i}^{u})\}_{i=1}^{n_{u}}$. We call $\widetilde{\mathbf{y}}_{i}^{u}$ as a `soft' pseudo label (i.e., an output prediction vector) and $\hat{y}_{i}^{u}$ as a `hard' pseudo label for the $i$th unlabeled target image, and they are given as follows:
\begin{equation}
\widetilde{\mathbf{y}}_{i}^{u}=[p(y=1\mid\mathbf{x}_{i}^{u}),p(y=2\mid\mathbf{x}_{i}^{u}),...,p(y=K\mid\mathbf{x}_{i}^{u})]^{T},
\label{soft_label}
\end{equation}
\begin{equation}
\hat{y}_{i}^{u}=\underset{k\in\{ 1, 2, ..., K\}}{\text{argmax}}p(y=k\mid\mathbf{x}_{i}^{u}).
\label{hard_label}
\end{equation}
In the above equations, $p(y=k\mid\mathbf{x}_{i}^{u})$ is the output probability of the $k$th class and $K$ denotes the number of classes. We empirically found that adopting the entire pseudo labels for training is not helpful and even degrades the performance. Our speculation regarding this problem is that training data whose pseudo labels are incorrect may degrade the accuracy, and thus acquiring pseudo labels with high reliability is a very important issue. Based on this observation, we propose a selective pseudo labeling approach that utilizes restricted amounts of pseudo labels by focusing on their reliabilities.

The key idea of our selective pseudo labeling approach is illustrated in Fig. \ref{fig:proposed1}. As depicted in the figure, deep features which lead to correct pseudo labels are closely located with those of labeled target images in the feature space. For each class, let $\mathbf{f}(\mathbf{x}_{i}^{t})$ be the feature of the $i$th labeled target image whose label is $k$, and $\mathbf{f}(\mathbf{x}_{j}^{u})$ be the feature of the $j$th unlabeled target image whose hard pseudo label is $k$ (i.e., $\hat{y}_{j}^{u}=k$). Here, we drop the categorical index $k$ for notational convenience. For the $j$th unlabeled sample, we define its feature distance $d_{j}$ as follows:
\begin{eqnarray}
d_{j}=\frac{1}{n_{t}^{\prime}}\sum_{i=1}^{n_{t}^{\prime}} \left \| \mathbf{f}(\mathbf{x}_{i}^{t})-\mathbf{f}(\mathbf{x}_{j}^{u})\right \|_{1},
\label{eqn:d}
\end{eqnarray}
where $\left \| \cdot \right \|_{1}$ denotes the $\textit{l}$1-norm function and $n_{t}^{\prime}$ indicates the number of labeled target images for each class. In our experiments, one or three target images are given for each class, i.e., $n_{t}^{\prime}$=1 (1-shot) or $n_{t}^{\prime}$=3 (3-shot). The feature distance $d_{j}$ becomes larger if the unlabeled target feature is located far from the labeled target features in the feature space and vice versa. Based on our assumption that $d_{j}$ is inversely proportional to the reliability, we sort the unlabeled features in an ascending order. This procedure is independently conducted for each class. After the sorting process, for each class, we assign pseudo labels to the first $n_{u}^{\prime}=\left \lceil r_{u}\frac{n_{u}}{K} \right \rceil$ samples. Here, $r_{u}$ is a hyper-parameter that adjust the ratio of selecting pseudo labels, and we set $r_{u}$ to $0.2$ as default. Through these procedures, we obtain the pseudo labeled target image set $\hat{\mathcal{D}}_{u}^{*}=\{(\mathbf{x}_{i}^{u}, \widetilde{\mathbf{y}}_{i}^{u}, \hat{y}_{i}^{u})\}_{i\in \mathcal{I}^{u}}$, where $\mathcal{I}^{u}$ indicates the index set of selected pseudo labels. The overall procedure of our selective pseudo labeling approach is illustrated in Fig. \ref{fig:proposed2}.

In Table \ref{tab:pseudo}, the reliabilities of selected pseudo labels are compared with those of baseline pseudo labels without applying the selective pseudo labeling approach. Here, it is worth noting that the numerical values in Table \ref{tab:pseudo} are not the final accuracy of the image classifier, but the ratio of correct pseudo labels in terms of percentage. For various adaptation scenarios in Table \ref{tab:pseudo}, our proposed selective approach consistently enhances the reliabilities of pseudo labels. In particular, its effectiveness becomes prominent when applied for adaptive scenarios with a large domain gap such as Clipart to Sketch (C to S). This indicates that the proposed selective pseudo labeling approach is fairly effective for challenging scenarios as well.

\begin{table}[!t]
\centering
\caption{Reliability of pseudo labels in terms of accuracy (\%) on the LSDAC dataset \cite{peng2019moment}. Before $\rightarrow$ After applying the selective pseudo labeling approach. Note that each measurement in this table is not the final accuracy, but the correctness of pseudo labels.}
\begin{tabular}{c|c c | c c}
\Xhline{4\arrayrulewidth}
\multirow{2}{*}{Net} & \multicolumn{2}{c|}{Clipart to Sketch (C to S)} & \multicolumn{2}{c}{Painintg to Real (P to R)}\\
& 1-shot & 3-shot & 1-shot & 3-shot\\
\hline
AlexNet & 35.2$\rightarrow$\textbf{61.6} & 41.0$\rightarrow$\textbf{64.8} & 57.7$\rightarrow$\textbf{83.8} & 60.7$\rightarrow$\textbf{85.8}\\
VGG-16 & 51.2$\rightarrow$\textbf{72.5} & 54.6$\rightarrow$\textbf{76.4} & 72.2$\rightarrow$\textbf{88.6} & 75.0$\rightarrow$\textbf{92.3}\\
\Xhline{4\arrayrulewidth}
\end{tabular}
\label{tab:pseudo}
\end{table}

\subsection{Label Noise-Robust Learning via Progressive Self-Training}
\label{sec:2B}
The final stage of our proposed method is to conduct SSDA along with the pseudo labels that are obtained by the previous stage. Although the pseudo labels are carefully determined via the selective approach, they are not completely reliable since the pseudo labels are noisy. Based on our observation that pseudo labels are inevitably noisy, we propose a label noise-robust learning approach, which is motivated by the joint optimization framework for learning with noisy labels \cite{tanaka2018joint}.

Given the set of unlabeled target images with pseudo labels ($\hat{\mathcal{D}}_{u}^{*}$), we implement the supervised loss function as follows:
\begin{equation}
\mathcal{L}_{pl}=\mathbb{E}_{(\mathbf{x}, \widetilde{\mathbf{y}})\in\hat{\mathcal{D}}_{u}^{*}} \mathcal{L}_{ce}(\mathbf{\mathbf{p}(\mathbf{x}), \widetilde{\mathbf{y}}}),
\label{L_pl}
\end{equation}
%\begin{equation}
%\mathcal{L}_{pl}=\frac{1}{\left | \mathcal{I}^{u}\right |} \sum_{i\in\mathcal{I}^{u}} \mathcal{L}_{ce}(\mathbf{\widetilde{\mathbf{y}}}_{i}^{u}, \mathbf{p}(\mathbf{x}_{i}^{u})),
%\label{L_pl}
%\end{equation}
where $\mathcal{L}_{ce}(\cdot, \cdot)$ is the standard cross entropy loss function. Note that $\mathbf{\widetilde{\mathbf{y}}}$ is a fixed pseudo label and $\mathbf{p}(\mathbf{x})$ is a variable output prediction during updating the network. In a similar way to \cite{tanaka2018joint}, the set of pseudo labels $\{\mathbf{\widetilde{\mathbf{y}}}_{i}^{u}\}_{i\in\mathcal{I}^{u}}$ is updated by forward passing operations using the updated network with a momentum of 0.9 after every validation phase. By means of this alternating learning process, the network and the set of pseudo labels are progressively updated. This procedure that jointly updates the network and the pseudo labels is continued until the validation accuracy is converged. We call this learning process as `progressive self-training' since the network is progressively optimized along with the pseudo labels.

The overall training is conducted in conjunction with the baseline SSDA method, which is the minimax entropy-based approach \cite{saito2019mme}. By letting $\mathcal{L}_{F}$ and $\mathcal{L}_{C}$ be the loss functions for the feature extractor and the classifier, respectively, the overall training objective functions are given as follows:
\begin{equation}
\mathcal{L}_{F}=\mathcal{L}_{l}+\mathcal{L}_{pl}+\lambda H,
\label{L_total1}
\end{equation}
\begin{equation}
\mathcal{L}_{C}=\mathcal{L}_{l}+\mathcal{L}_{pl}-\lambda H,
\label{L_total2}
\end{equation}
\begin{equation}
\mathcal{L}_{l}=\mathbb{E}_{(\mathbf{x}, y)\in\mathcal{D}_{s}, \mathcal{D}_{t}} \mathcal{L}_{ce}(\mathbf{p}(\mathbf{x}), y),
\label{L_l}
\end{equation}
\begin{equation}
H=-\mathbb{E}_{\mathbf{x}\in\mathcal{D}_{u}} \sum_{i=1}^{K}p(y=i|\mathbf{x})\,\text{log}(p(y=i|\mathbf{x})).
\label{H}
\end{equation}
In the above equations, $\mathcal{L}_{l}$ is the standard cross entropy loss for labeled source and target images and $H$ indicates the entropy \cite{saito2019mme} for unlabeled target images. The standard Stochastic Gradient Descent (SGD) algorithm is used for training on the loss functions. The hyper-parameter $\lambda$ is set to 0.1 for all experiments. The overall training procedure is summarized in Algorithm 1.

\section{Experiments}
\label{sec:exp}
\subsection{Datasets}
We used three representative benchmark datasets for experiments. \textbf{LSDAC}\cite{peng2019moment} is a benchmark dataset for large-scale domain adaptation, which involves 6 domains with 345 classes. To make a fair comparison with previous methods, we followed the settings in \cite{saito2019mme}, which addresses 7 adaptation scenarios from 4 domains (Real, Clipart, Painting, and Sketch) with 126 classes. \textbf{Office-Home}\cite{venkateswara2017deep} contains 4 domains (Real, Clipart, Art, and Product) with 65 classes and we conducted evaluations on 12 adaptation scenarios, which involve all possible scenarios. \textbf{Office}\cite{saenko2010adapting} involves 3 domains (Amazon, Webcam, and DSLR) with 31 classes and we evaluated on 2 scenarios, which are Webcam to Amazon and DSLR to Amazon. Since the domain disparities between Webcam and DSLR are negligible, we considered two domain adaptation scenarios that involve large domain shifts and sufficient amount of training data.

\subsection{Experimental Setups}
For each adaptation scenario, one or three examples per class are used as labeled target training data, and we denote these two settings as `1-shot' and `3-shot', respectively. For fair comparison, we used labeled target image sets, which are reported in \cite{saito2019mme}. The rest of unlabeled target images and all labeled source images were used for training. To verify the effectiveness of the proposed method across various network models, we conducted comparative evaluations on 6 backbone architectures. To be specific, we employed AlexNet\cite{alex2012alexnet}, VGG-16\cite{vggnet}, and ResNet-34\cite{resnet} as the primary network models. Further results on other models beyond the three architectures are reported in Sec. \ref{sec:ablation}.

\renewcommand{\algorithmicrequire}{\textbf{Input:}}
\renewcommand{\algorithmicensure}{\textbf{Output:}}

\begin{algorithm}[!t]
\caption{Semi-supervised Domain Adaptation with the Label Noise-Robust Learning Approach}
\begin{algorithmic}
\REQUIRE $\mathcal{D}_{s}$, $\mathcal{D}_{t}$, $\mathcal{D}_{u}$, $\hat{\mathcal{D}}_{u}^{*}$, $\hat{\theta}_{F}$, $\hat{\theta}_{C}$
\ENSURE $\theta_{F}^{*}$, $\theta_{C}^{*}$
\STATE $t_{iter} \leftarrow 1$, $t_{max}\leftarrow50\text{k}$, $t_{val}\leftarrow0.5\text{k}$
\WHILE{$t_{iter} < t_{max}$ \textbf{and} not converged}
\STATE{update $\hat{\theta}_{F}$ by SGD on $\mathcal{L}_{F}$ in Eq. (\ref{L_total1})}
\STATE{update $\hat{\theta}_{C}$ by SGD on $\mathcal{L}_{C}$ in Eq. (\ref{L_total2})}
\IF{$t_{iter}\text{\%}t_{val}=0$}
\STATE{update $\{\mathbf{\widetilde{\mathbf{y}}}_{i}^{u}\}_{i\in\mathcal{I}^{u}}$ with a momentum of 0.9}
\ENDIF
\STATE $t_{iter} \leftarrow t_{iter}+1$
\ENDWHILE
\STATE{$\theta_{F}^{*}\leftarrow \hat{\theta_{F}}$, $\theta_{C}^{*}\leftarrow \hat{\theta_{C}}$}
\RETURN $\theta_{F}^{*}, \theta_{C}^{*}$
\end{algorithmic}
\end{algorithm}

\begin{table*}[!t]
\centering
\caption{Quantitative evaluation results on LSDAC dataset in terms of accuracy (\%).}
\begin{tabular}{c|c|c@{\,\;}c@{\,\;}c@{\,\;}c@{\,\;}c@{\,\;}c@{\,\;}c@{\,\;}c@{\,\;}c@{\,\;}c@{\,\;}c@{\,\;}c@{\,\;}c@{\,\;}c | c@{\,\;}c}
\Xhline{4\arrayrulewidth}
\multirow{2}{*}{Net} & \multirow{2}{*}{Method} & \multicolumn{2}{c}{R to C} & \multicolumn{2}{c}{R to P} & \multicolumn{2}{c}{P to C} & \multicolumn{2}{c}{C to S} & \multicolumn{2}{c}{S to P} & \multicolumn{2}{c}{R to S} & \multicolumn{2}{c|}{P to R} & \multicolumn{2}{c}{MEAN}\\
& & 1-shot & 3-shot & 1-shot & 3-shot & 1-shot & 3-shot & 1-shot & 3-shot & 1-shot & 3-shot & 1-shot & 3-shot & 1-shot & 3-shot & 1-shot & 3-shot\\
\hline
\multirow{7}{*}{AlexNet} & S+T & 43.3 & 47.1 & 42.4 & 45.0 & 40.1 & 44.9 & 33.6 & 36.4 & 35.7 & 38.4 & 29.1 & 33.3 & 55.8 & 58.7 & 40.0 & 43.4\\
& DANN & 43.3 & 46.1 & 41.6 & 43.8 & 39.1 & 41.0 & 35.9 & 36.5 & 36.9 & 38.9 & 32.5 & 33.4 & 53.6 & 57.3 & 40.4 & 42.4\\
& ADR & 43.1 & 46.2 & 41.4 & 44.4 & 39.3 & 43.6 & 32.8 & 36.4 & 33.1 & 38.9 & 29.1 & 32.4 & 55.9 & 57.3 & 39.2 & 42.7\\
& CDAN & 46.3 & 46.8 & 45.7 & 45.0 & 38.3 & 42.3 & 27.5 & 29.5 & 30.2 & 33.7 & 28.8 & 31.3 & 56.7 & 58.7 & 39.1 & 41.0\\
& ENT & 37.0 & 45.5 & 35.6 & 42.6 & 26.8 & 40.4 & 18.9 & 31.1 & 15.1 & 29.6 & 18.0 & 29.6 & 52.2 & 60.0 & 29.1 & 39.8\\
& MME & 48.9 & 55.6 & 48.0 & 49.0 & 46.7 & 51.7 & 36.3 & 39.4 & 39.4 & 43.0 & 33.3 & 37.9 & 56.8 & 60.7 & 44.2 & 48.2\\
& Proposed & \textbf{54.2} & \textbf{58.3} & \textbf{48.8} & \textbf{51.7} & \textbf{49.0} & \textbf{55.1} & \textbf{38.9} & \textbf{43.5} & \textbf{44.7} & \textbf{48.4} & \textbf{37.5} & \textbf{41.2} & \textbf{60.2} & \textbf{63.3} & \textbf{47.6} & \textbf{51.6}\\
\hline\hline
\multirow{7}{*}{VGG-16} & S+T & 49.0 & 52.3 & 55.4 & 56.7 & 47.7 & 51.0 & 43.9 & 48.5 & 50.8 & 55.1 & 37.9 & 45.0 & 69.0 & 71.7 & 50.5 & 54.3\\
& DANN & 43.9 & 56.8 & 42.0 & 57.5 & 37.3 & 49.2 & 46.7 & 48.2 & 51.9 & 55.6 & 30.2 & 45.6 & 65.8 & 70.1 & 45.4 & 54.7\\
& ADR & 48.3 & 50.2 & 54.6 & 56.1 & 47.3 & 51.5 & 44.0 & 49.0 & 50.7 & 53.5 & 38.6 & 44.7 & 67.6 & 70.9 & 50.2 & 53.7\\
& CDAN & 57.8 & 58.1 & 57.8 & 59.1 & 51.0 & 57.4 & 42.5 & 47.2 & 51.2 & 54.5 & 42.6 & 49.3 & 71.7 & 74.6 & 53.5 & 57.2\\
& ENT & 39.6 & 50.3 & 43.9 & 54.6 & 26.4 & 47.4 & 27.0 & 41.9 & 29.1 & 51.0 & 19.3 & 39.7 & 68.2 & 72.5 & 36.2 & 51.1\\
& MME & 60.6 & 64.1 & 63.3 & 63.5 & 57.0 & 60.7 & 50.9 & 55.4 & 60.5 & 60.9 & 50.2 & 54.8 & 72.2 & \textbf{75.3} & 59.2 & 62.1 \\
& Proposed & \textbf{64.5} & \textbf{68.0} & \textbf{63.7} & \textbf{64.9} & \textbf{60.5} & \textbf{64.4} & \textbf{53.7} & \textbf{57.4} & \textbf{62.5} & \textbf{63.4} & \textbf{52.7} & \textbf{57.5} & \textbf{73.0} & 74.9 & \textbf{61.5} & \textbf{64.4}\\
\hline\hline
\multirow{7}{*}{ResNet-34} & S+T & 55.6 & 60.0 & 60.6 & 62.2 & 56.8 & 59.4 & 50.8 & 55.0 & 56.0 & 59.5 & 46.3 & 50.1 & 71.8 & 73.9 & 56.9 & 60.0\\
& DANN & 58.2 & 59.8 & 61.4 & 62.8 & 56.3 & 59.6 & 52.8 & 55.4 & 57.4 & 59.9 & 52.2 & 54.9 & 70.3 & 72.2 & 58.4 & 60.7\\
& ADR & 57.1 & 60.7 & 61.3 & 61.9 & 57.0 & 60.7 & 51.0 & 54.4 & 56.0 & 59.9 & 49.0 & 51.1 & 72.0 & 74.2 & 57.6 & 60.4\\
& CDAN & 65.0 & 69.0 & 64.9 & 67.3 & 63.7 & 68.4 & 53.1 & 57.8 & 63.4 & 65.3 & 54.5 & 59.0 & 73.2 & 78.5 & 62.5 & 66.5\\
& ENT & 65.2 & 71.0 & 65.9 & 69.2 & 65.4 & 71.1 & 54.6 & 60.0 & 59.7 & 62.1 & 52.1 & 61.1 & 75.0 & 78.6 & 62.6 & 67.6\\
& MME & 70.0 & 72.2 & 67.7 & 69.7 & 69.0 & 71.7 & 56.3 & 61.8 & 64.8 & 66.8 & 61.0 & 61.9 & 76.1 & 78.5 & 66.4 & 68.9\\
& Proposed & \textbf{72.4} & \textbf{73.9} & \textbf{69.4} & \textbf{71.5} & \textbf{71.6} & \textbf{73.9} & \textbf{61.7} & \textbf{63.3} & \textbf{66.7} & \textbf{69.0} & \textbf{62.5} & \textbf{65.1} & \textbf{78.8} & \textbf{80.4} & \textbf{69.0} & \textbf{71.0}\\
\Xhline{4\arrayrulewidth}
\end{tabular}
\label{tab:domainnet}
\end{table*}

All experiments in this paper are implemented in PyTorch \cite{pytorch} by using an NVIDIA TITAN X GPU (Pascal architecture). For training baseline models (i.e., the first stage of our method), we followed the setups reported in \cite{saito2019mme}. The self-training phase using the selected pseudo labels (Eq. (\ref{L_total1}), (\ref{L_total2})) is resumed from the baseline models until the validation accuracy is converged. Learning rates are initialized before resuming the training process, and are decayed according to the annealing strategy proposed in \cite{ganin2015dann}. For comparative evaluations, we report the quantitative evaluation results of the following 6 previous methods. \textbf{S+T}\cite{chen2019closer, ranjan2017l2} is a method that trains a network with supervisions on labeled source and target images without using unlabeled target images. \textbf{DANN}\cite{ganin2015dann}, \textbf{ADR}\cite{saito2017adversarial}, \textbf{CDAN}\cite{long2018cdan}, and \textbf{ENT}\cite{grandvalet2005semi} are unsupervised domain adaptation methods, which are trained with additional supervisions on labeled target images. \textbf{MME}\cite{saito2019mme} is our baseline method that is specialized to the SSDA scheme.

\subsection{Experimental Results and Analysis}
The quantitative evaluation results on the LSDAC dataset is reported in Table \ref{tab:domainnet}. For the 7 adaptation scenarios, the proposed method outperforms other previous methods except only one case (P to R with VGG16). It is worth noting that our method achieves significant performance increasements over the baseline method when the domain gap is large (e.g., S to P and R to S adaptation scenarios). This implies that our method is particularly robust to challenging conditions. Another empirical observation is that the 1-shot accuracies of our method are competitive to or even higher than those of 3-shot accuracies of other previous methods. This indicates that our method requires less target labels than other methods for the same performance. Therefore, our method can be used as an alternative to collecting labeled images in the target domain. This advantage of our method would be very useful for image classification tasks involving a large number of classes since the expense of annotating labels is proportional to the number of classes. The evaluation results on the Office-Home and the Office datasets are reported in Table \ref{tab:office}. Our method shows better performances than other methods in terms of average accuracies.

Overall, the strength of our method can be summarized as the following three major aspects. First, our proposed method outperforms other previous methods across various datasets and network architectures. This indicates that our method can be broadly adopted to various SSDA scenarios, not limited to a certain dataset or network. Second, our method achieves considerable performance enhancements over the previous methods, especially for large-scale domain adaptation datasets such as the LSDAC dataset. Third, our method is particularly robust to challenging domain adaptation scenarios (e.g., S to P and R to S adaptation scenarios in the LSDAC dataset), implying that our proposed method can be used for enhancing performance for more difficult adaptation conditions involving large domain shifts.

\begin{table}[!t]
\centering
\caption{Quantitative evaluation results on Office-Home and Office datasets in terms of accuracy (\%). Each measurement is a mean accuracy averaged over all adaptation scenarios in each dataset (12 and 2 scenarios for Office-Home and Office datasets, respectively).}
\begin{tabular}{c|c|c@{\quad}c@{\quad}c@{\quad}c@{\quad}}
\Xhline{4\arrayrulewidth}
\multirow{2}{*}{Net} & \multirow{2}{*}{Method} & \multicolumn{2}{c}{Office-Home} & \multicolumn{2}{c}{Office}\\
& & 1-shot & 3-shot & 1-shot & 3-shot\\
\hline
\multirow{7}{*}{AlexNet} & S+T & 44.1 & 50.0 & 50.2 & 61.8\\
& DANN & 45.1 & 50.3 & 55.8 & 64.8\\
& ADR & 44.5 & 49.5 & 50.6 & 61.3\\
& CDAN & 41.2 & 46.2 & 49.4 & 60.8\\
& ENT & 38.8 & 50.9 & 48.1 & 65.1\\
& MME & 49.2 & 55.2 & 56.5 & 67.6\\
& Proposed & \textbf{50.3} & \textbf{55.3} & \textbf{59.0} & \textbf{69.8}\\
\hline\hline
\multirow{7}{*}{VGG-16} & S+T & 57.4 & 62.9 & 68.7 & 73.3\\
& DANN & 60.0 & 63.9 & 69.8 & 75.0\\
& ADR & 57.4 & 63.0 & 69.4 & 73.7\\
& CDAN & 55.8 & 61.8 & 65.9 & 72.9\\
& ENT & 51.6 & 64.8 & 70.6 & 75.3\\
& MME & 62.7 & 67.6 & 73.4 & 77.0\\
& Proposed & \textbf{63.9} & \textbf{68.6} & \textbf{76.4} & \textbf{78.1}\\
\Xhline{4\arrayrulewidth}
\end{tabular}
\label{tab:office}
\end{table}

\subsection{Ablation Studies and Further Analysis}
\label{sec:ablation}
To verify the effectiveness of each module in our method, we conducted ablation studies. The ablation studies were done on the two adaptation scenarios in the LSDAC dataset, which are C to S involving a large domain gap and P to R with a relatively small domain gap. In Table \ref{tab:ablation1}, the accuracies depending on the ratio of selecting pseudo labels ($r_{u}$ in Sec. \ref{sec:2A}) are reported. The results in Table \ref{tab:ablation1} demonstrate that the accuracy has a tendency to be maximized when $r_{u}$ is around 0.2. Meanwhile, the performance is degraded if the magnitude of $r_{u}$ is larger or smaller than $0.2$. This indicates that selecting moderate amounts of pseudo labels is encouraged. If $r_{u}=1.0$, the entire pseudo labels are adopted for self-training. Thus, this setup corresponds to the training strategy without applying the selective pseudo labeling stage in Sec. \ref{sec:2A}. By comparing the results of $r_{u}=1.0$ with those of $r_{u}=0.2$, it can be validated that the selective pseudo labeling stage obviously enhances the accuracy. In addition, this result demonstrates our initial assumption that employing a restricted number of pseudo labels with high reliability leads to better performance than adopting the entire pseudo labels. On the other hand, adopting too small amount of pseudo labels leads to relatively low accuracies. This empirical observation implies that a moderate number of pseudo labels are desirable for self-training. Based on these analysis and empirical studies, we set the default value of $r_{u}$ to 0.2 for all experiments.

The second ablation study is to investigate the effectiveness of the label noise-robust learning approach (Sec. \ref{sec:2B}). To this end, we compared our method with a vanilla learning approach by using hard pseudo labels in Eq. (\ref{hard_label}) without applying the progressive updating scheme. The comparative results are presented in Table \ref{tab:ablation2} and it can be confirmed that the performance of the label noise-robust learning approach is better than that of the vanilla learning approach. This indicates that the proposed learning approach can effectively prevent incorrect pseudo labels from misleading the network during the training phase.

Lastly, we conducted comparative evaluations on three additional backbone architectures to verify the robustness of the proposed method across various network models. We adopted ResNet-101\cite{resnet}, DenseNet-121\cite{densenet} to test on deeper network models. In addition, we employed MobileNet-v2 \cite{mobilenet} to confirm the performance on a light-weight network model. The evaluation results on the three models are reported in Table \ref{tab:backbone}. Our proposed method surpasses other previous SSDA methods including the baseline method. This consistency of performance enhancements indicates that the proposed method can be broadly applied for SSDA without demanding any preference on a certain network architecture. To train a single DA scenario, it took around 4 to 6 hours until convergence. The computational time for testing is dependent on the backbone architecture, and the measurements for the 6 network models are reported in Table \ref{tab:time}.

\begin{table}[!t]
\centering
\caption{Accuracy variations on $r_{u}$ in Sec. \ref{sec:2A} using AlexNet.}
\begin{tabular}{c | c c c c c}
\Xhline{4\arrayrulewidth}
$r_{u}$ & 0.01 & 0.05 & 0.20 & 0.50 & 1.00\\
\hline
\multicolumn{6}{c}{1-shot}\\
\hline
C to S & 36.7 & 37.6 & \textbf{38.9} & 38.7 & 37.4\\
P to R & 57.5 & 59.3 & \textbf{60.2} & 59.8 & 59.6\\
\hline
\multicolumn{6}{c}{3-shot}\\
\hline
C to S & 41.7 & 42.8 & \textbf{43.5} & 43.1 & 42.4\\
P to R & 60.5 & 62.1 & \textbf{63.3} & 62.9 & 62.0\\
\Xhline{4\arrayrulewidth}
\end{tabular}
\label{tab:ablation1}
\end{table}

\begin{table}[!t]
\centering
\caption{Ablation study on applying the label noise-robust learning approach in Sec. \ref{sec:2B} using AlexNet.}
\begin{tabular}{c|c c c c}
\Xhline{4\arrayrulewidth}
\multirow{2}{*}{Whether applied} & \multicolumn{2}{c}{C to S} & \multicolumn{2}{c}{P to R}\\
& 1-shot & 3-shot & 1-shot & 3-shot\\
\hline
Yes & \textbf{38.9} & \textbf{43.5} & \textbf{60.2} & \textbf{63.3}\\
No & 37.7 & 42.1 & 58.3 & 61.9\\
\Xhline{4\arrayrulewidth}
\end{tabular}
\label{tab:ablation2}
\end{table}

\begin{table}[!t]
\centering
\caption{Further evaluation results on various network architectures. Each measurement is a mean accuracy (\%) averaged over the 7 adaptation scenarios in LSDAC dataset.}
\begin{tabular}{c|c@{\;\;}cc@{\;\;}cc@{\;\;}c}
\Xhline{4\arrayrulewidth}
\multirow{2}{*}{Method} & \multicolumn{2}{c}{ResNet-101} & \multicolumn{2}{c}{DenseNet-121} & \multicolumn{2}{c}{MobileNet-v2}\\
& 1-shot & 3-shot & 1-shot & 3-shot & 1-shot & 3-shot\\
\hline
S+T & 55.9 & 59.1 & 58.6 & 61.8 & 51.3 & 54.6\\
ENT & 62.1 & 67.0 & 62.2 & 69.7 & 53.7 & 61.7\\
MME & 66.3 & 68.4 & 68.3 & 70.5 & 60.9 & 64.3\\
Proposed & \textbf{68.0} & \textbf{69.2} & \textbf{70.4} & \textbf{72.1} & \textbf{63.5} & \textbf{66.3}\\
\Xhline{4\arrayrulewidth}
\end{tabular}
\label{tab:backbone}
\end{table}

\begin{table}[!t]
\centering
\caption{Computation time required for testing an image.}
\begin{tabular}{c|c|c}
\Xhline{4\arrayrulewidth}
AlexNet & VGG-16 & ResNet-34\\
\hline
1.84 ms (544 FPS) & 2.41 ms (414 FPS) & 1.82 ms (550 FPS)\\
\hline
\hline
ResNet-101 & DenseNet-121 & MobileNet-v2\\
\hline
2.46 ms (406 FPS) & 1.83 ms (547 FPS) & 1.82 ms (550 FPS)\\
\Xhline{4\arrayrulewidth}
\end{tabular}
\label{tab:time}
\end{table}

\section{Conclusion}
\label{sec:conclusion}
In this paper, we have introduced a novel semi-supervised domain adaptation method for image classification. The major idea of our method is to exploit the labeled target images to find out reliable pseudo labels for the unlabeled target images. In addition, based on the observation that the set of pseudo labels may contain incorrect labels, a learning approach that is robust to noisy labels is applied. Experimental results on the three representative domain adaptation datasets show that our method outperforms other methods, especially for the challenging adaptation scenarios involving large domain shifts. For the three primary backbone architectures (AlexNet\cite{alex2012alexnet}, VGG-16\cite{vggnet}, ResNet-34\cite{resnet}), the SSDA method outperforms the previous state-of-the-art method by 2.7\%, 0.9\%, and 2.2\% for LSDAC\cite{peng2019moment}, Office-Home\cite{venkateswara2017deep}, and Office\cite{saenko2010adapting} datasets, respectively. \color{black} Though we validated the proposed method on image classification only, we expect that our method could be further expanded to other computer vision tasks such as domain adaptive object detection\cite{chen2018domain} and semantic segmentation\cite{hoffman2017cycada} in the future.

\section*{Acknowledgement}
We thank the anonymous reviewers for their valuable comments. This work was supported by Institute of Information \& Communications Technology Planning \& Evaluation(IITP) grant funded by the Korea government(MSIT) (No. 2019-0-00524, Development of precise content identification technology based on relationship analysis for maritime vessel/structure).

\bibliographystyle{IEEEtran}
\bibliography{bibtex}

\end{document}